%% file: bare_jrnl_new_sample4.tex
\documentclass[lettersize,journal]{IEEEtran}
\usepackage{amsmath,amsfonts}
\usepackage{algorithmic}
\usepackage{algorithm}
\usepackage{array}
\usepackage[caption=false,font=normalsize,labelfont=sf,textfont=sf]{subfig}
\usepackage{textcomp}
\usepackage{stfloats}
\usepackage{url}
\usepackage{verbatim}
\usepackage{graphicx}
\usepackage{cite}
\input{math_commands2.tex}

\usepackage{hyperref}
\usepackage{xcolor}
\usepackage{booktabs}
\hypersetup{
    colorlinks=true,
    linkcolor=blue,
    filecolor=magenta,      
    urlcolor=blue,
}

\usepackage{multirow}
\usepackage{bbm}
\usepackage{url}

\usepackage{caption}
\usepackage{boldline}
\usepackage{amsthm}
\theoremstyle{plain}
\newtheorem{theorem}{Theorem}
\newtheorem*{IMSVD*}{IMSVD Theorem}

\begin{document}

\title{Information-Maximized Soft Variable Discretization for Self-Supervised Image Representation Learning}

\author{Chuang Niu,~\IEEEmembership{Member,~IEEE,}
        Wenjun Xia,~\IEEEmembership{Member,~IEEE,}
        Hongming Shan,~\IEEEmembership{Senior Member,~IEEE,}
        and~Ge~Wang,~\IEEEmembership{Fellow,~IEEE}
\thanks{C. Niu, W. Xia, and G. Wang are with Department of Biomedical Engineering, Center for Biotechnology and Interdisciplinary Studies, Rensselaer Polytechnic Institute, Troy, NY USA, 12180. E-mail: niuc@rpi.edu; xiaw4@rpi.edu; wangg6@rpi.edu.}
\thanks{H. Shan is with the Institute of Science and Technology for Brain-inspired Intelligence and  MOE Frontiers Center for Brain Science and Key Laboratory of Computational Neuroscience and Brain-Inspired Intelligence, Fudan University, Shanghai, 200433, China, and also with the Shanghai Center for Brain Science and Brain-inspired Technology, Shanghai 201210, China. E-mail: hmshan@fudan.edu.cn.}
\thanks{This paper has supplementary downloadable material available at http://ieeexplore.ieee.org, provided by the author. The material includes the equation derivation, theorem proof, and more implementation details. Contact niuc@rpi.edu and wangg6@rpi.edu for further questions about this work.}
\thanks{Manuscript received Aug. 15, 2024.}}

\markboth{}%
{Shell \MakeLowercase{\textit{et al.}}: A Sample Article Using IEEEtran.cls for IEEE Journals}


\maketitle

\begin{abstract}
Self-supervised learning (SSL) has emerged as a crucial technique in image processing, encoding, and understanding, especially for developing today's vision foundation models that utilize large-scale datasets without annotations to enhance various downstream tasks.
This study introduces a novel SSL approach, Information-Maximized Soft Variable Discretization (IMSVD), for image representation learning.
Specifically, IMSVD softly discretizes each variable in the latent space, enabling the estimation of their probability distributions over training batches and allowing the learning process to be directly guided by information measures.
Motivated by the MultiView assumption, we propose an information-theoretic objective function to learn transform-invariant, non-travail, and redundancy-minimized representation features.
We then derive a joint-cross entropy loss function for self-supervised image representation learning, which theoretically enjoys superiority over the existing methods in reducing feature redundancy.
Notably, our non-contrastive IMSVD method statistically performs contrastive learning. 
Extensive experimental results demonstrate the effectiveness of IMSVD on various downstream tasks in terms of both accuracy and efficiency. Thanks to our variable discretization, the embedding features optimized by IMSVD offer unique explainability at the variable level.
IMSVD has the potential to be adapted to other learning paradigms.
Our code is publicly available at \url{https://github.com/niuchuangnn/IMSVD}.
\end{abstract}


\section{Introduction}
\IEEEPARstart{S}{elf}-supervised learning (SSL) has emerged as a critical learning paradigm in computer vision through unleashing the power of large-scale datasets and foundation models~\cite{bommasani2021opportunities}. Without annotations, SSL leverages intrinsic structures and relationships within data to optimize models and empowers various applications, such as image denoising~\cite{tip-noise, noise2sim}, super-resolution~\cite{tip2}, clustering~\cite{niu2020gatcluster, spice}, representation learning~\cite{simclr, tip1, moco, bardes2021vicreg}, and understanding~\cite{tip-object, tip-seg}. In this study, we focus on self-supervised image representation learning and its downstream applications in image understanding tasks including classification, object detection, and instance segmentation.

\begin{figure*}[bt]
    \centering
    \includegraphics[width=1\textwidth]{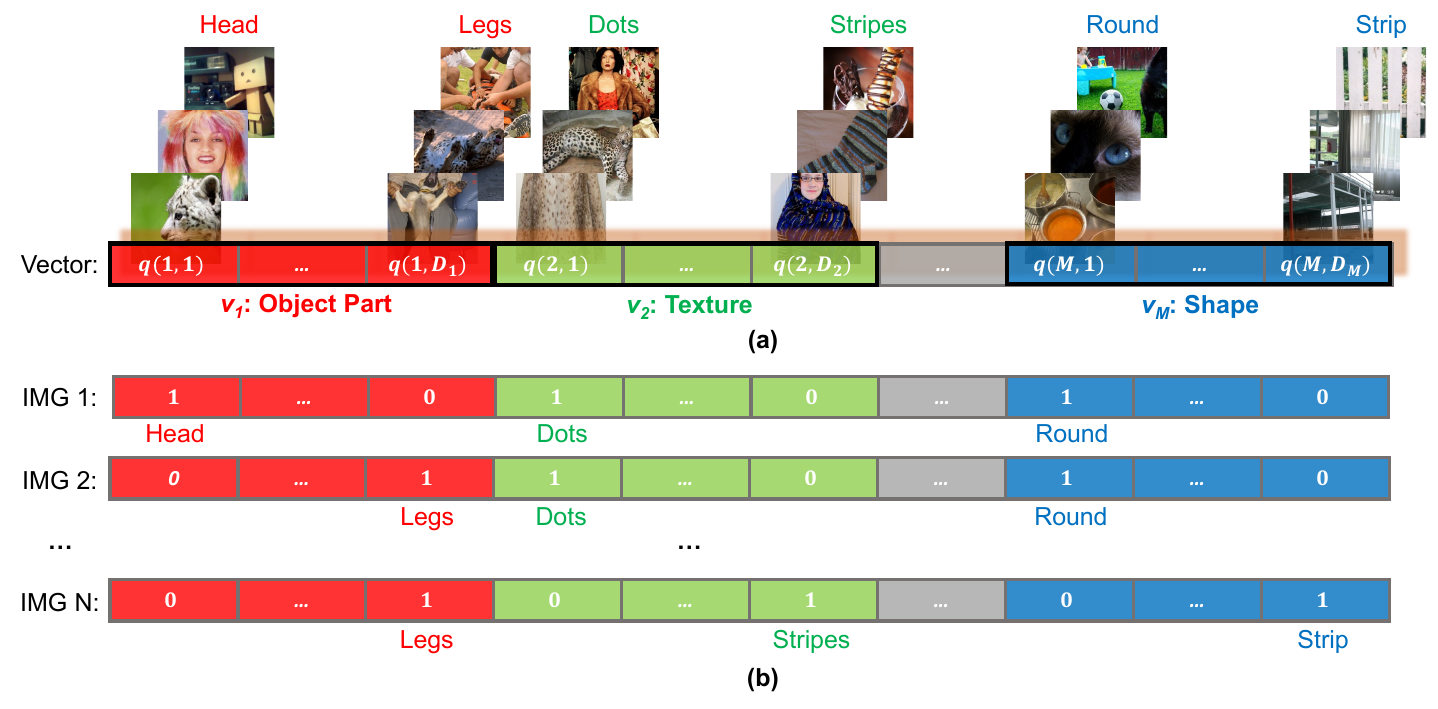}
    \caption{Illustration of discrete variables for encoding images. 
    (a) The feature vector is statistically optimized to be a set of discrete variables ($v_1$, ..., $v_M$) as shown in different colors. Different variables are associated with diverse attributes; \textit{e.g.}, $v_1, v_2, v_M$ represent object part, text, and shape, respectively. Each variable $v_m$ is quantized with a set of discrete values represented by a one-hot vector $\vq(m, :)$.
    The example images are selected with the practically optimized vectors in the same way as described in Sec.~\ref{sec_vis}.
    (b) Specific images are encoded with different combinations of the one-hot vectors; \textit{e.g.}, IMG-1 is encoded with $v_1 = [1, 0, \cdots, 0]$ representing the head part, $v_2 = [1, 0, \cdots, 0]$ representing the dots texture, and $v_M = [1, 0, \cdots, 0]$ representing the round shape.
    }
    \label{fig_vector}
\end{figure*}

A popular SSL approach is to drive different views of the same instance close to each other in the latent space~\cite{dosovitskiy2014discriminative}.
Its effectiveness relies on the MultiView assumption that the shared information between different views is sufficient for downstream tasks~\cite{shwartz2023compress}. 
For SSL, various augmentations of the same image are regarded as different views.
This assumption is practical when the downstream task is not affected by such data augmentations~\cite{NEURIPS2020_4c2e5eaa}.
Based on the MultiView assumption, the representation features are expected to be transform-invariant, non-collapsed, and redundancy-minimized. It is straightforward to induce transform-invariant features by maximizing the similarity or minimizing the distance between the embedding vectors of different views.
However, simply maximizing similarity or reducing distance tends to produce collapsed solutions; \textit{i.e.}, all samples are mapped into a single point (total collapse) or a subspace (dimensional collapse) in the whole latent space~\cite{ozsoy2022self}.
Thus, a primary challenge for SSL is preventing collapses while learning informative representations.
Contrastive and non-contrastive methods are two main categories to overcome the collapses for SSL. 
Recent efforts have been made to study the relationship between contrastive and non-contrastive methods, which is a crucial direction to help unify current SSL methods and develop more advanced SSL approaches.
To reduce the feature redundancy, current methods~\cite{michael2018on, bardes2021vicreg} proposed to minimize the pair-wise linear correlation among feature variables. However, the redundancy of non-linear correlation and the total correlation among all feature variables are not considered.

Information theory has been playing an important role in understanding and designing SSL methods~\cite{shwartz2023compress}.
For example, it is demonstrated that optimizing the InfoNCE loss for contrastive methods maximizes the lower bound of mutual information between different views~\cite{henaff2020data}.
Assuming the embedding features are Gaussian with some simplifications, the non-contrastive method, Barlow Twins, is connected to the information bottleneck principle~\cite{michael2018on} that explains informational features being learned~\cite{zbontar2021barlow}.
Current SSL methods are often analyzed based on these approximate connections to information theory.
A natural question is: why not directly apply the information measures to optimizing the representation learning model other than explaining them?
The well-known difficulty to this question is the direct computing of information measurements in a high-dimensional and continuous latent space mapped by deep neural networks~\cite{shwartz2023compress}.

To overcome the above challenges, we propose Information-Maximized Soft Variable Discretization (IMSVD) as a novel SSL approach directly grounded in information theory.
Specifically, each variable in the feature vector is softly quantized into a set of discrete units using the softmax function. 
In contrast to ``hard'' discretization~\cite{shwartz2017opening}, the soft discretization operator is differentiable and can be easily integrated into end-to-end optimization.
For intuitive understanding, we associate the discrete feature variables to attribute learning~\cite{russakovsky2010attribute} that represents objects with a set of discrete attributes. 
As illustrated in Fig.~\ref{fig_vector}, the embedding vector consists of multiple variables (in different colors) describing diverse types of attributes; \textit{e.g.}, $v_1$, $v_2$, and $v_M$ represent object configuration, texture, and shape, respectively. Each variable is discretized into discrete units encoded by a soft one-hot vector, each of which represents a specific attribute; \textit{e.g.}, the units of variable $v_2$ represent different textural patterns, such as dots, stripes, etc.
Note that we cannot ensure that all practically learned units correspond to well-defined attributes, but some discrete units are indeed explainable and correspond to well-known attributes in our empirical analysis; see Sec.~\ref{sec_vis}.
Given a batch of samples, the probability distribution of each latent variable over its discrete units can be estimated. Thus, information measures defined on these probability distributions can be directly computed for optimization and analysis.

Given soft variable discretization, we propose a principled objective function with information-theoretic measures to learn transform-invariant, non-collapsed, and redundancy-minimized representation features in the SSL setting. Then, an equivalent joint cross-entropy loss function is derived to maximize the information of latent variables.
In contrast to the Gumbel-softmax~\cite{jang2016categorical,maddison2017the} that selects a small relaxation/temperature hyperparameter for one-hot encoding, we show that optimizing the joint entropy loss will drive latent variables to be one-hot encoded without manually tuning the softmax temperature.
Theoretically and practically, we demonstrate that our non-contrastive IMSVD method optimizes image representation features to be discrete, transform-invariant, non-collapsed, redundancy-minimized, and discriminative.

The contributions of this paper are as follows.
(1) We propose a novel SSL approach that softly discretizes feature variables making their probability distributions estimable so that the information-theoretic measures can be directly computed and applied for model optimization. 
(2) We present a novel information-theoretic objective for SSL and a simplified yet effective cross-joint entropy loss function that directly calculates and maximizes the information of latent variables. In particular, this loss function can minimize any form of dependency between feature variables beyond the linear correlation regularized by current methods~\cite{zbontar2021barlow, bardes2021vicreg,ozsoy2022self} so that using a shallower projector and a lower-dimension embedding vector achieves even better results with less computational costs than baseline methods.
(3) Our theoretical analysis ensures that IMSVD optimizes the representation features to be discrete, transform-invariant, non-collapsed, and redundancy-minimized, which are verified by the corresponding empirical results.
Although IMSVD is a non-contrastive method without explicitly using negative samples, we show that it statistically performs contrastive learning, providing another aspect to understanding contrastive and non-contrastive SSL methods.
(4) Extensive experimental results demonstrate the effectiveness and efficiency of IMSVD on common downstream tasks, including linear classification, k-NN classifications, object detection, and instance segmentation, achieving either better or competitive performance relative to the state-of-the-art SSL methods.

\section{Related Work}

\subsection{Discretization in Deep Learning} 
Discretization has empowered various machine learning methods for improved results~\cite{kotsiantis2006discretization,Garca2013ASO}.
To understand deep neural networks through information theory, the discretization technique was used to divide the continuous neuron activations into discrete bins so that the probability distributions of discrete variables and the mutual information between them can be calculated~\cite{shwartz2017opening,michael2018on, goldfeld2019estimating}, which we refer them to ``hard" discretization as hard thresholds are used for binning.
However, the ``hard" discretization is not differentiable, and thus it cannot be used for optimizing models.
Recently, discretization has been successfully applied in generative models including discrete variational autoencoders (dVAE)~\cite{ramesh2021zero,bao2021beit} and vector-quantized autoencoders (VQ-VAE)~\cite{van2017neural,esser2021taming, tip2}.
To address the non-differentiable problem, dVAE~\cite{ramesh2021zero} uses Gumbel-softmax relaxation~\cite{jang2016categorical,maddison2017the} during training, which selects a small relaxation temperature to approximate one-hot vectors for discretization.
VQ-VAE~\cite{van2017neural} leverages a set of learnable vectors to generate discrete index features by the nearest matching between the encoder prediction and codebook vectors. VQ-VAE~\cite{van2017neural} addresses the non-differentiable problem with the straight-through strategy that directly copies the gradients of decoder inputs as the approximated gradients of encoder outputs.

In contrast, the variables in the IMSVD feature vector are optimized to be one-hot through the proposed loss function instead of manually setting a small relaxation value as in dVAE~\cite{ramesh2021zero}. On the other hand, IMSVD can accurately compute the gradients and does not need additional codebook vectors for VQ-VAE~\cite{van2017neural}.

\subsection{Self-Supervised Learning}

\textbf{Contrastive methods}~\cite{simclr,moco,wu2018unsupervised, NEURIPS2020_4c2e5eaa, ye2019unsupervised, dwibedi2021little} explicitly push away different instances while pulling together different augmentations of the same instance in the latent space.
Usually, large batch sizes are required, such as for SimCLR~\cite{simclr}.
MoCo~\cite{moco} uses a momentum updating technique and a memory bank~\cite{wu2018unsupervised} to store a large number of features as negative samples so that small batch sizes can be used.
All these methods are based on InfoNCE loss that estimates the lower bound on mutual information between different views~\cite{van2018representation}.

\textbf{Clustering Methods}~\cite{caron2018deep, caron2019unsupervised, asano2019self, yan2020clusterfit, huang2019unsupervised, zhuang2019local, gidaris2021obow,spice} leverage conventional clustering algorithms for SSL.
DeepCluster~\cite{caron2018deep} iteratively performs k-means and updates a neural network using the cluster assignments. To avoid collapsed solutions, random samples are selected for the empty cluster to compute the centroid.
SELA~\cite{asano2019self} leverages the Sinkhorn-Knopp algorithm to iteratively perform clustering and optimize the clustering networks with the assigned cluster labels on the fly.
SwAV~\cite{caron2020unsupervised} alternatively computes the cluster assignment of one view and optimizes the network to predict the same assignment for other views of the same sample.

\begin{figure*}[t]
    \centering
    \includegraphics[width=1\textwidth]{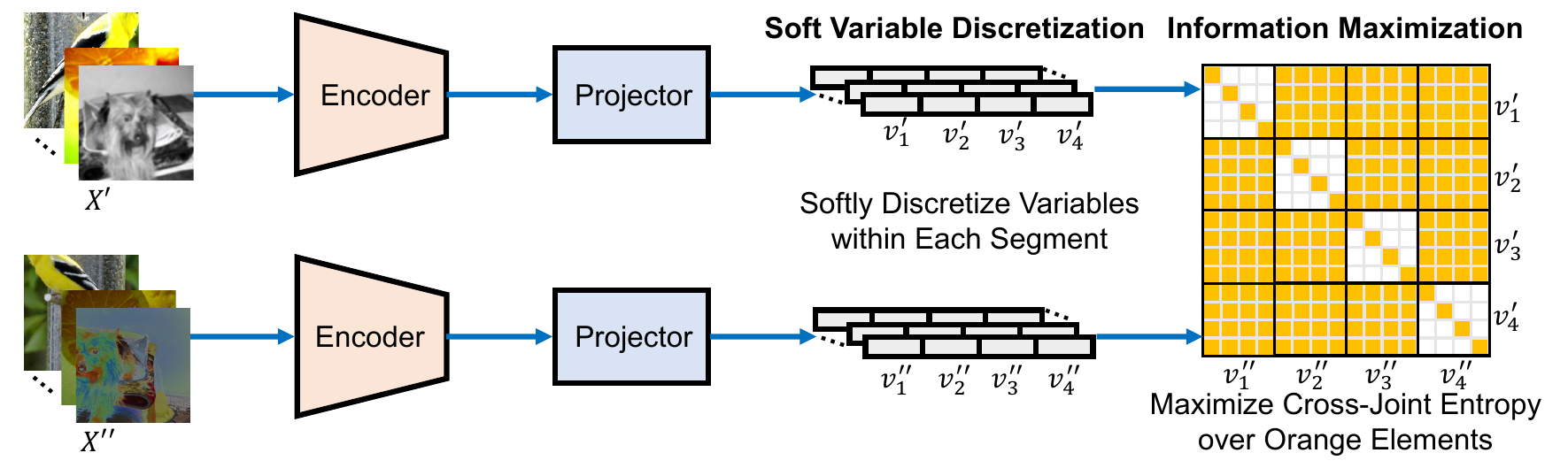}
    \caption{SSL framework through IMSVD optimized with the joint entropy loss. For illustration purposes, the embedding feature vector only consists of four variables and each variable is discretized into four units.}
    \label{fig_ssl}
\end{figure*}

\textbf{Non-contrastive methods} use an asymmetric architecture design or covariance-based regulation to avoid collapsed features without explicitly using negative samples.
BYOL~\cite{richemond2020byol} and SimSiam~\cite{Chen_2021_CVPR} use asymmetric architectures and stop gradient techniques to avoid trivial solutions. 
Recently, covariance regulation methods~\cite{ermolov2021whitening,zbontar2021barlow,bardes2021vicreg} propose to train a simple twin architecture using covariance matrix based loss functions without needing any asymmetric techniques.
WMSE~\cite{ermolov2021whitening} minimizes the MSE distance between different views and enforces the self-covariance matrix to be an identity matrix.
Barlow Twins~\cite{zbontar2021barlow} optimizes the cross-covariance matrix towards an identity matrix.
VICReg~\cite{bardes2021vicreg} integrates three loss terms including invariance, variance, and covariance.
CorInfoMax~\cite{ozsoy2022self} uses log-determinant mutual information for SSL, which reflects linear
dependence and is equivalent to the Shannon information measures when embedding vectors are Gaussian distributed.

\textbf{Understanding SSL approaches} has recently attracted much attention as it is important to guide the development of more advanced SSL approaches. For example, theoretical analyses~\cite {understand, zhang2021does, richemond2020byol, tian2021understanding} and visual explanation methods~\cite{tip1} have been conducted to reveal why contrastive methods and asymmetric designs can avoid collapsed solutions and learn meaningful features.
In~\cite{balestriero2022contrastive,garrido2023on}, the contrastive and covariance-based non-contrastive methods are bridged from different theoretical perspectives, showing both differences and similarities between these two families of SSL methods.

Our IMSVD method is closely related to covariance-based non-contrastive methods using neither asymmetric design nor negative samples.
Different from the linear decorrelation regulation in current methods, IMSVD presents a way to overcome the difficulty of directly computing information in the latent space and enables a novel information-theoretic loss function that minimizes redundancy beyond linear correlation.
Theoretically, our non-contrastive IMSVD statistically performs contrastive learning.

\section{Approach}
\label{sec_method}

\subsection{MultiView SSL Framework}

MultiView SSL framework assumes that the shared information between different views is sufficient for downstream tasks~\cite{shwartz2023compress}, and thus, drives different views of the same instance close to each other in the latent space~\cite{dosovitskiy2014discriminative}.
For self-supervised image representation learning, various augmentations of the same image are regarded as different views.
In this study, we adopt a twin architecture~\cite{zbontar2021barlow,bardes2021vicreg}, where the same network weights are shared between two branches, as shown in Fig.~\ref{fig_ssl}.
During training, input images $\sX=\{\vx_i\}_{i=1}^N$ are randomly transformed to two augmentation sets $\sX'=\{\vx'_i\}_{i=1}^N$ and $\sX''=\{\vx''_i\}_{i=1}^N$, where $N$ is the batch size.
A common transformation distribution, which covers random crops combined with color distortions, the same as that in~\cite{bardes2021vicreg}, is used to generate training samples.
Then, the two sets of distorted images $\sX'$ and $\sX''$ are respectively fed to two branches, each of which consists of an encoder $f(\cdot; \vtheta_f)$ and a projector $g(\cdot; \vtheta_g)$, where $\vtheta_f$ and $\vtheta_g$ respectively denote the parameters of the encoder and projector to be optimized.
Given an image $\vx$, the output of projector is $\vz = g(f(\vx_i; \vtheta_f); \vtheta_g) \in \R^D$, where $D$ is the feature dimension.
During training, we use $\vz'_i$ and $\vz''_i$ to denote the embedding vectors of two augmentations $\vx'_i$ and $\vx''_i$ from the same image. 
Note that the presented method is not limited to this twin architecture, which could be extended to the two branches with different parameters, heterogeneous networks, and different input modalities.

\subsection{Soft Variable Discretization}
Here we propose to softly discretize feature variables and estimate both individual probability distributions and joint probability distributions.
As illustrated in Fig.~\ref{fig_vector}, each variable is discretized into a set of discrete units encoded by a one-hot sub-vector, and the whole embedding vector consists of multiple such one-hot sub-vectors.
We reformat the projector output as $\vz(m, d), d=1,\cdots, D_m, m=1,\cdots, M$, where $M$ is the number of variables, $D_m$ is the dimension of the $m^{th}$ variable.
In this study, all variables have the same number of discrete units, \textit{i.e.}, $\forall m, D_m = D_M$, and the dimension of the whole feature vector is $D = D_M \times M$. In principle, different variables can be discretized into different numbers of discrete units given prior knowledge.
Then, the softly discretized variables are calculated with the softmax function:
\begin{equation}
    \vq(m, d) = \frac{\exp(\vz(m, d))}{\sum_{k=1}^{D_M} \exp(\vz(m, k))},
    \label{eq_prob}
\end{equation}
where $\vq(m, d),m=1,\cdots, M, d=1,\cdots, D_m$ is defined as the discrete feature vector.
Note that $\forall m, \vq_i(m, :)$ will be optimized to be a ``hard'' one-hot vector with our proposed objective function; see Sec.~\ref{sec_analysis}.
Therefore, a sub-vector $\vq(m, :)$ represents a feature variable with $D_m$ different discrete values; \textit{e.g.}, $\vq(m, d)=1$ means that the $m^{th}$ feature variable takes its $d^{th}$ discrete value. 
Sometimes, we also denote the discrete feature vector as $\vq = [v_1, v_2, \cdots, v_M]$ for convenience, where $v_m$ is the $m^{th}$ discrete feature variable.

Given a sufficient number of independent samples, the probability distribution of each variable over its limited number of discrete values can be estimated as:
\begin{equation}
    \vp(m, d) = \frac{1}{N} \sum_{i=1}^{N} \vq_i(m, d),
\end{equation}
where $\vp_i(m, d)$ denotes the probability of the $m^{th}$ variable taking its $d^{th}$ value.
Then, the joint distribution of multiple variables can be estimated as:
\begin{equation}
    \mP(\vq^r; \vd^r) = \frac{1}{N} \sum_{i=1}^{N} \vq_i(m_1, d_1)\vq_i(m_2, d_2)\cdots \vq_i(m_r, d_r),
    \label{eq_jp}
\end{equation}
where $\vq^r = [v_{m_1},v_{m_2}, \cdots, v_{m_r}]$ denotes $m^{th}_1, m^{th}_2, \cdots, m^{th}_r$ discrete feature variables, $\vd^r = [d_1, d_2, \cdots, d_r]$ denotes $d_1^{th}, d^{th}_2, \cdots, d_r^{th}$ discrete values, $\mP(\vq^r; \vd^r)$ denotes the joint probability of the variables $v_{m_1},v_{m_2}, \cdots, v_{m_r}$ taking their $d_1^{th}, d^{th}_2, \cdots, d_r^{th}$ values, respectively and simultaneously.

\subsection{Information-theoretic Objective for SSL}
\label{sec_objective}

With the MultiView SSL framework, the embedding features are expected to be transform-invariant, non-collapsed, and redundancy-minimized. For this purpose, we propose to maximize the following objective:
\begin{equation}
    \label{eq_obj1}
    \begin{split}
            \max_{\vtheta_f, \vtheta_g} &\frac{1}{N}\sum_{i=1}^{N} <\vq'_i, \vq''_i> + \frac{\lambda}{2} (\overline{S}^{(1)}(\vq') + \overline{S}^{(1)}(\vq'')) \\
            &- \frac{\beta}{2} (\overline{C}^{(r)}(\vq') + \overline{C}^{(r)}(\vq'')),
    \end{split}
\end{equation}
where the first term is to learn transform-invariant features by maximizing the feature similarity between different views, $<\cdot, \cdot>$ is a kernel function to measure the similarity between two feature vectors; the second term is to maximize the average entropy of all individual variables in the embedding feature vector to avoid the collapsed solution that all samples have the same embedding feature; the third term is to minimize the average total correlation of $r$-variable subsets so that different variables represent diverse features; $\lambda$ and $\beta$ are the coefficients. As in \cite{watanabe1960information}, the average entropy $\overline{S}^{(r)}(\vq)$ and total correlation $\overline{C}^{(r)}(\vq)$ of $r$-variable subsets are defined as:
\begin{equation}
    \label{eq_srbar}
    \overline{S}^{(r)}(\vq) = - \sum_{\vq^r \in \vq} S^{(r)}(\vq^r) \Bigg/ \binom{M}{r},
\end{equation}
\begin{equation}
    \label{eq_sr}
    S^{(r)}(\vq^r) = - \sum_{d_1=1}^{D_{m_1}}\sum_{d_2=1}^{D_{m_2}}\cdots \sum_{d_r=1}^{D_{m_r}} \mP(\vq^r; \vd^r)\log \mP(\vq^r; \vd^r),
\end{equation}
where $S^{(r)}(\vq^r)$ is the joint entropy of feature variables $\vq^r$, which can be calculated in Eq.~\eqref{eq_sr}, the average entropy $\overline{S}^{(r)}(\vq)$ is averaging the joint entropy of all $r$-variable subsets, and the number of $r$-variable subsets out of total $M$ variables is $\binom{M}{r}$.
Given Eqs.~\eqref{eq_srbar} and \eqref{eq_sr}, when $r=1$ we have:
\begin{equation}
    \overline{S}^{(1)} = \frac{1}{M} \sum_{m=1}^{M} S(v_m)=-\frac{1}{M} \sum_{m=1}^{M} \sum_{d=1}^{D_m} \vp(m, d) \log \vp(m, d),
\end{equation}
The total correlation $\overline{C}^{(r)}(\vq)$ of $r$-variable subsets is defined as:
\begin{equation}
    \label{eq_cr}
    \overline{C}^{(r)}(\vq) = r \overline{S}^{(1)}(\vq) -  \overline{S}^{(r)}(\vq),
\end{equation}
when $r=M$, $\overline{C}^{(r)}(\vq) = \sum_{m=1}^{M}S(v_m) - S(v_1, v_2, \cdots, v_m)$ is the total correlation among all $M$ feature variables; please refer to \cite{watanabe1960information} for more explanations. 
In Eq.~\eqref{eq_obj1}, the average entropy of all individual variables, \textit{i.e.}, $\overline{S}^{(1)}(\vq)$, will be maximized, and thus, minimizing the average total correlation, $\overline{C}^{(r)}(\vq)$, is equivalent to maximizing the average entropy of all $r$-variable subsets, $\overline{S}^{(r)}(\vq)$,  given the Eq.~\eqref{eq_cr}.
Then, the SSL objective function in Eq.~\eqref{eq_obj1} can be reformulated as:
\begin{equation}
    \label{eq_obj2}
    \begin{split}
            \max_{\vtheta_f, \vtheta_g} &\frac{1}{N}\sum_{i=1}^{N} <\vq'_i, \vq''_i> + \frac{\lambda}{2} (\overline{S}^{(1)}(\vq') + \overline{S}^{(1)}(\vq'')) \\
            &+ \frac{\beta}{2} (\overline{S}^{(r)}(\vq')+\overline{S}^{(r)}(\vq'')),
    \end{split}
\end{equation}

\subsection{Cross-Joint Entropy Loss for SSL}
\label{sec_method_loss}

In this study, we set $r=2$ to minimize the average total correlation in Eqs.~\eqref{eq_obj1} and~\eqref{eq_obj2}, considering the high-computation cost of estimating the multivariate probability distributions in Eq.~\eqref{eq_jp}. Nevertheless, the redundancy reduction is still stronger than the pair-wise linear correlation reduction in existing methods~\cite{zbontar2021barlow, bardes2021vicreg}; see Sec.~\ref{sec_analysis} for detailed analysis.
Furthermore, we define the cross-joint discrete probability distribution between two variables, $(v_{m_1},v_{m_2})$, as:
\begin{equation}
    \mP^c(m_1, m_2; d_1, d_2) = \frac{1}{N} \sum_{i=1}^{N} \vq'_i(m_1, d_1)\vq''_i(m_2, d_2).
    \label{eq_cjp}
\end{equation}
Since the embedding features are optimized to be transform-invariant; \textit{i.e.}, $\vq'_i = \vq''_i$ in the Eq.~\eqref{eq_obj2}, the cross-joint probability will be equal to the self-joint entropy; \textit{i.e.}, $\mP^c(m_1, m_2; d_1, d_2) = \mP(m_1, m_2; d_1, d_2)$. 
Then, the cross-joint entropy loss function can be derived from Eq.~\eqref{eq_obj2} as follows (see Appendix-I for details):
\begin{equation}
    \label{eq_loss}
    \begin{split}
    &L = - \frac{1}{NM}\sum_{i,m}^{N,M} \log {\vq'(m,:)}^T_i \vq''_i(m,:) + \frac{\lambda}{M^2} \sum_{m_1,m_2}^{M}\sum_{d_1,d_2}^{D_M} \\
    &(1 - \1^\mathrm{m_1=m_2}_\mathrm{d_1\ne d_2}) \mP^c(m_1, m_2; d_1, d_2) \log\mP^c(m_1, m_2; d_1, d_2), 
    \end{split}
\end{equation}
where the logarithmic inner product is used to measure the feature similarity. We also explored the cross-entropy function to measure the similarity between two embedding vectors, \textit{i.e.}, $-\frac{1}{NM} \sum_{i=1}^{N}\sum_{m=1}^M\sum_{d=1}^{D_M} \vq'_i(m, d)\log(\vq''_i(m, d))$. However, by using the cross-entropy the performance would be degraded, as reported in Sec.~\ref{sec_empirical}. By default, we set $\lambda = \beta = 1$, which is in principle neither too small nor too large for a good balance. $\1^\mathrm{m_1=m_2}_\mathrm{d_1\ne d_2}$ is an indicator function that equals to 1 if $m_1 = m_2$ and $d_1 \ne d_2$; otherwise, it equals to 0. The empirical cross-joint distribution can be denoted by a block matrix as shown in Fig.~\ref{fig_ssl}, where $(1 - \1_\mathrm{m_1=m_2, d_1\ne d_2})$ means picking up the diagonal elements of the diagonal blocks and all elements of the off-diagonal blocks, as indicated by the orange elements. 
Thus, minimizing the second loss term is maximizing the cross-joint entropy over the orange elements.
Our proposed method can be efficiently implemented, with a PyTorch-style pseudo-code in Appendix-IV.

\subsection{Theoretical Analysis}
\label{sec_analysis}

First of all, we prove a theorem (the complete proof can be found in Appendix-II) as follows:
\begin{theorem}[IMSVD Theorem]
\label{theorem1}
    If the cross-joint entropy loss function in Eq.~\eqref{eq_loss} is minimized, we have: for $\forall i, m, d, \vq'_i(m,d) = \vq''_i(m,d)$ are one-hot vectors,  $\vp(m,d)=\frac{1}{D_M}$, $\forall m_1, m_2, d_1, d_2, m_1 \ne m_2, \mP(m_1, m_2; d_1, d_2) = \mP^c(m_1, m_2; d_1, d_2) = \frac{1}{(D_M)^2}$, and the mutual information between any two variables is zero.
\end{theorem}

Given the results of the IMSVD theorem, we analyze the properties of the embedding features optimized by our IMSVD method as follows.

\textbf{Transform invariance}:
The solution that $\forall i, m, \vq'_i(m, :)$ = $\vq''_i(m, :)$ means that the learned features invariant to transformations, and each variable is one-hot encoded meaning that variables are optimized to be ``hard'' discretized.

\textbf{Minimum redundancy}:
The solution that the mutual information between any two different variables is zero means the redundancy between any two variables is minimized.
From another perspective, $\forall m_1, m_2, d_1, d_2, m_1 \ne m_2$, we have $\mP(m_1, m_2, d_1, d_2) = \frac{1}{(D_M)^2} = \vp(m_1, d_1) \vp(m_2, d_2)$, and thus any two variables in the feature vector are independent.
In contrast to the linear correlation regulation methods~\cite{zbontar2021barlow, bardes2021vicreg}, our information-theoretic objective reduces any form of redundancy between two variables more than the linear correlation.
This property is further validated by our empirical results, which show that using a shorter embedding vector in our method achieved better performance than state-of-the-art baseline methods; see Sec.~\ref{sec_empirical}.

\textbf{Non-collapsed features}:
The solution $\vp(m,d)=\frac{1}{D_M}$ means the samples are evenly distributed over discrete units for all variables, avoiding the total collapse features that all samples have the same embedding feature.
Furthermore, dimensional collapse can be avoided with minimized redundancy as analyzed in~\cite{hua2021feature}.

\textbf{Discriminative encoding}:
Contrastive learning or instance discrimination has proven very effective for representation learning by maximizing the similarity between different transformations of the same instance while discriminating the reference from other instances.
It is underlined that IMSVD is consistent with contrastive learning and discriminating instances in a novel way.
Specifically, since $\mP(m_1, m_2, d_1, d_2) = \frac{1}{(D_M)^2}$, any two discrete variables in the optimal feature vector will encode $(D_M)^2$ different samples.
In our default settings $D_M=80$ (see Sec.~\ref{sec_implement} for details), each pair of variables can represent $6,400$ different samples.
Since the number of all possible embeddings is larger ($2,048 < 6,400$) than the batch size, it will be enforced to encode different instances with different embeddings. 
Like contrastive learning, our discrete embedding vectors are optimized to be discriminative among different samples.
The difference lies in that contrastive learning differentiates instances by pushing the reference away from its negative instances through pairwise cosine similarity, while IMSVD statically assigns instances with discriminative discrete features in an information-maximized manner.
A formal demonstration of our IMSVD as dimensional conservative learning can be found in Appendix-III.

In Sec.~\ref{sec_vis}, the individual embedding vector and the empirical cross-joint probability matrix optimized by IMSVD on the ImageNet dataset are visualized, showing that the empirical results are consistent with the theoretical analysis.
In summary, the discrete features optimized with the information-theoretic objective are discrete, transform-invariant, non-collapsed, redundancy-minimized, and discriminative.

\section{Implementation Details}
\label{sec_implement}
Covariance regulation methods, VICReg~\cite{bardes2021vicreg} and Barlow Twins (BT) ~\cite{zbontar2021barlow}, are used as the baseline because only these methods optimize a simple twin architecture with a pure SSL loss function without needing any asymmetric techniques or negative samples.
For a fair comparison with the baseline methods, we closely followed their implementation settings to train our IMSVD models.
Specifically, the standard ResNet-50 backbone~\cite{He_2016_CVPR} was used as the encoder that outputs a representation vector of 2,048 units in the same training settings, including the data augmentation (random cropping, horizontal flip, color jittering, grayscale, Gaussian blur, solarization, with the same parameters in VICReg), the optimizer of LARS~\cite{you2017large, goyal2017accurate} with a weight decay of $10^{-6}$ and the learning rate of $lr=batch\_size/256 \times base\_lr$, and the cosine decay schedule~\cite{loshchilov2016sgdr} from 0 with 10 warmup epochs towards the final value of 0.002. The base learning rate $base\_lr$ was set to 0.6 in our study. By default, we used a two-layer MLP projector (8,192-8,160), the number of segments $M=102$, the segment dimension $D_M=80$, and $D=D_M \times M=8,160$ (similar to the feature dimension of $8,192$ used by VICReg and BT).
The results were respectively analyzed for different feature dimensions, depths of projectors, batch sizes, sub-vector dimensions, and numbers of training epochs.
The effectiveness of the single extra hyperparameter $D_M$ of IMSVD was evaluated as well.
The SSL models were trained on the 1,000-classes ImageNet dataset without labels and evaluated in various downstream tasks.

\textbf{ImageNet Linear classification:}
For all evaluation experiments on ImageNet linear classification, we followed the standard procedure that a linear classifier was trained on top of the frozen backbone of a ResNet-50 pre-trained with IMSVD. The SGD optimizer was used with a learning
rate of 0.02, a cosine decay, a weight decay of $10^{-6}$, a batch size of 256, and 100 training epochs. 
In the training stage, the images were augmented by the composition of random cropping and resizing of ratio 0.2 to 1.0 for size 224$\times$224, and random horizontal flips. 
In the testing stage, the images were simply cropped from the image center and resized to $224 \times 224$.

\textbf{Object detection and instance segmentation:}
Mask R-CNN ~\cite{he2017mask} with the C-4 backbone was trained on the COCO 2017 train split and tested on the validation set. 
We used a learning rate of 0.1 and kept the other parameters the same as in the 1 schedule in Detectron2~\cite{wu2019detectron2}.

\textbf{Transfer Learning Linear classification:}
We followed the exact settings from PIRL~\cite{misra2020self} in evaluating linear classifiers on the Places-205 and VOC07 datasets. For Places-205, a linear classifier was trained using the SGD optimizer for 14 epochs with a learning rate of 0.01 reduced by a factor of 10 at epochs 5 and 10, a weight decay of $5\times 10^{-4}$, and a momentum of 0.9.
For VOC2007 dataset, we trained SVM classifiers, where the $C$ values were computed using cross-validation.

IMSVD models were distributively trained on four nodes, each of which has 2$\times$ 20 core 2.5 GHz Intel Xeon Gold 6248 and 8$\times$ NVIDIA Tesla V100 GPU each with 32 GiB HBM.
The codes are publicly available at \url{https://github.com/niuchuangnn/IMSVD}.

\section{Experiments and Results}

\subsection{Evaluation Results in Different Tasks}

\begin{table*}[htp]
  \renewcommand\tabcolsep{32pt}
 \caption{\textbf{Linear classification on ImageNet}. Top-1 accuracy (in \%) is reported. The best results are highlighted in \textbf{bold} while the second best results are \underline{underlined}. * means the results were obtained with 300 training epochs.}
  \centering
  \begin{tabular}{lccccc}
  \hlineB{2}
  Epochs                                & 100               & 200                     & 400               & 800               & 1000 \\
  \hline
  SimCLR\cite{simclr}                   & 66.5              & 68.3                    & 69.8              & 70.4              & - \\
  MoCoV2\cite{moco}                     & 67.4              & 69.9                    & 71.0              & 72.2              & - \\  
  BYOL\cite{richemond2020byol}          & 66.5              & 70.6                    & \textbf{73.2}     & \textbf{74.3}     & \textbf{74.3} \\ 
  SwAV\cite{caron2020unsupervised}      & 66.5              & 69.1                    & 70.7              & 71.8              & - \\ 
  SimSiam\cite{Chen_2021_CVPR}          & 68.1              & 70.0                    & 70.8              & 71.3              & - \\
  DISSL\cite{dubois2022improving}       & 68.9              & -                       & -                 & -                 & - \\
  CorInfoMax\cite{ozsoy2022self}        & \underline{69.1}  & \underline{71.4}        & -                 & -                 & - \\
  BT\cite{zbontar2021barlow}            & -                 & \underline{71.4*}  & -                 & -                 & 73.2 \\
  VICReg\cite{bardes2021vicreg}         & 68.7              & 71.2*              & -                 & -                 & 73.2 \\
  IMSVD                                 & \textbf{69.4}     & \textbf{71.8}           &\textbf{73.2}      & \underline{73.4}  & \underline{73.6} \\
  \hlineB{2}
  \end{tabular}
  \label{tab:compare}
\end{table*}

\begin{table*}[htp]
  \renewcommand\tabcolsep{12pt}
 \caption{\textbf{KNN classification}. Top-1 accuracy with 20 and 200 nearest neighbors are reported. The best results are highlighted in \textbf{bold}.}
  \centering
  \begin{tabular}{lcccccccc}
  \hlineB{2}
  Methods  & NPID\cite{wu2018unsupervised} & LA\cite{zhuang2019local} & PCL\cite{li2021prototypical} & BYOL\cite{grill2020bootstrap} & SwAV\cite{caron2020unsupervised} & BT\cite{zbontar2021barlow} & VICReg\cite{bardes2021vicreg} & IMSVD \\
  \hline
  20-NN    & -    & -     & 54.5 & 66.7 & 65.7 & 64.8 & 64.5& \textbf{67.0} \\

  200-NN    & 46.5 & 49.4 & - & \textbf{64.9} & 62.7 & 62.9 & 62.9 & \textbf{64.9} \\
    \hlineB{2}
  \end{tabular}
  \label{tab:knn}
\end{table*}

\begin{table*}[htp]
  \renewcommand\tabcolsep{15pt}
 \caption{\textbf{Transfer Learning}. For object detection and instance segmentation tasks, SSL models pre-trained on ImageNet were used to initialize the backbone of the object detection and instance segmentation models on COCO. Mask R-CNN~\cite{he2017mask} with the C4 backbone variant~\cite{wu2019detectron2} was fine-tuned using the 1 schedule. AP metrics defined by COCO are reported here. For the linear classification task, Top-1 accuracy (in \%) for Places205~\cite{zhou2014learning} and mAP for VOC07~\cite{everingham2010pascal} are based on the frozen representations pre-trained on ImageNet.
 The best results are in \textbf{bold}.}
  \centering
  \begin{tabular}{ccccccccc}
  \hlineB{2}
     \multirow{2}{*}{Methods}      & \multicolumn{3}{c}{Object Detection} & \multicolumn{3}{c}{Instance Segmentation} & \multicolumn{2}{c}{Linear Classification} \\
                                   & AP$^{bb}$ & AP$^{bb}_{50}$ & AP$^{bb}_{75}$ & AP$^{mk}$ & AP$^{mk}_{50}$ & AP$^{mk}_{75}$ & VOC2007 & Places205 \\
    \hline
Sup. &  38.2 &58.2 &41.2 &33.3 &54.7 &35.2 &87.5&53.2 \\
\hline
MoCo-v2 & \textbf{39.3} &58.9 &42.5 & \textbf{34.4} &55.8 &36.5 &86.4 &51.8 \\
SwAV & 38.4 &58.6 &41.3 &33.8 &55.2 &35.9 &86.4 &52.8 \\
SimSiam & 39.2& \textbf{59.3}& 42.1& \textbf{34.4} & \textbf{56.0}& \textbf{36.7}&- &- \\
BT  &39.2 &59.0& 42.5 &34.3 & \textbf{56.0} &36.5 &86.2 &54.1 \\
IMSVD (Ours)  &  \textbf{39.3} & 59.1 & \textbf{42.6} & \textbf{34.4} & 55.8 & 36.6   &  \textbf{86.5} &  \textbf{54.8} \\
  \hlineB{2}
  \end{tabular}
  \label{tab:transfer}
\end{table*}

\subsubsection{Linear Classification on ImageNet.}
Linear probing is a common evaluation protocol that trains a linear classifier on top of frozen representations to evaluate the performance of SSL methods.
Being consistent with BT and VICReg, a ResNet-50 backbone was trained with the batch size of 2,048 on the training set of ImageNet, and the linear classification results in terms of Top-1 accuracies of different methods on the evaluation set are reported under different numbers of training epochs, as shown in Table \ref{tab:compare}.
IMSVD used a two-layer MLP projector (8,192-8,160) instead of three layers (8,192-8,192-8,192) for BT and VICReg.
The performance of IMSVD is on par with highly-engineered models, such as BYOL, which uses asymmetric techniques including an additional predictor and a momentum encoder.
Note that the results of the methods~\cite{caron2020unsupervised, gidaris2021obow} based on multi-crop/multi-positive techniques are not included in Table~\ref{tab:compare}. These techniques can usually boost performance further.
The comparative results show that IMSVD consistently outperforms BT and VICReg, where all these methods used a twin architecture without using negative pairs or any asymmetric techniques.

\subsubsection{KNN Classification on ImageNet.}
Another common protocol for evaluating representation learning methods is by K-Nearest-Neighbors (KNN) classification on ImageNet.
We followed the recent studies~\cite{wu2018unsupervised, zhuang2019local, caron2020unsupervised, bardes2021vicreg} that built KNN classifiers with the learned representations on the training set of ImageNet and evaluated the KNN classification results on the validation set of ImageNet.
The results with 20 and 200 nearest neighbors are reported in Table~\ref{tab:knn}, showing that IMSVD achieved the best performance among the comparison methods.
Since the KNN classifier determines the class of a sample by directly searching its nearest samples in the feature space, the representation features learned by IMSVD are more semantically similar among the nearest neighbors than those learned by other methods.
Thus, IMSVD has the potential superiority when applied to the downstream tasks of searching for the nearest neighbors.

\subsubsection{Transfer Learning.}
Transfer learning is a common way to evaluate SSL methods, including object detection, instance segmentation, and linear classification. Our results are reported in Table~\ref{tab:transfer}.
It is worth mentioning that different studies have varying setups for the object detection and instance segmentation tasks. Here we closely followed~\cite{zbontar2021barlow} selecting the same comparison methods in the same settings.
IMSVD performs on par with the current methods and better than BT on the object detection and segmentation tasks.
On the other hand, the linear classification results on VOC2007 and Places205 datasets show that IMSVD achieved better results than the compared methods.
Same as other SSL methods, IMSVD can effectively improve the downstream tasks in the transfer learning settings and achieves better results than the supervised pretraining counterparts in most cases.

\begin{table}[htp]
  \renewcommand\tabcolsep{2pt}
 \caption{\textbf{Running time and peak memory}. Comparison of different methods in terms of the running time over 100 epochs, the peak memory on a single GPU, and the top-1 accuracy (\%) on linear classification on top of the frozen representations.
 All models were distributively trained on 32 Tesla V100 GPUs.}
  \centering
  \begin{tabular}{cccc}
  \hlineB{2}
     Method         & Time/100epochs & Peak memory/GPU & Top-1 accuracy \\
    \hline
     SwAV                &  9h   & 9.5G & 71.8  \\
     BYOL                &  10h  & 14.6G & 74.3  \\
     Barlow Twins        &  12h  & 11.3G & 73.2  \\
     VICReg              &  11h  & 11.3G & 73.2  \\
     IMSVD               &  8.5h & 10.4G & 73.6  \\
    \hlineB{2}
  \end{tabular}
  \label{tab:run}
\end{table}

\subsubsection{Efficiency.}
In Table~\ref{tab:run}, the computational cost of IMSVD was evaluated and compared with other methods. All methods were run on 32 Tesla V100 GPUs.
These methods offer different trade-offs among running time, memory and performance.
SwAV with multi-crop and BYOL achieve better performance at the additional computational cost and memory usage. Barlow Twins and VICReg have balanced results with less memory than BYOL, but a slightly worse performance.
Compared with the most related Barlow Twins and VICReg methods, IMSVD cannot only reduce the running time and memory usage significantly but also improve performance. It is due to that IMSVD can use a shallower MLP head and a shorter embedding vector for a better performance as discussed in Sec.~\ref{sec_empirical}.
The computational cost of IMSVD will be significantly reduced further when using a ($\times$2) lower dimension for embeddings, and the performance is degraded very slightly, as discussed in Sec.~\ref{sec_empirical}.

All the above results demonstrate the effectiveness and superiority of IMSVD as a new SSL method principled by the information theory.
In the following subsections, the characteristics and superiority of IMSVD will be further discussed.

\subsection{Ablation Study}
\label{sec_empirical}
In this subsection, we comprehensively evaluate the proposed IMSVD method in various settings and compare it with other SSL methods if the corresponding results in the same or comparable settings were already reported. All the models were evaluated with linear classification on ImageNet.

\begin{table}[htp]
  \renewcommand\tabcolsep{15pt}
 \caption{\textbf{Batch Size.} Top-1 accuracy (in \%) results for linear classification on ImageNet were obtained based on ResNet50 with 100 pre-training epochs. The best results are highlighted in \textbf{bold}.}
  \centering
  \begin{tabular}{lcccc}
  \hlineB{2}
  Batch Size  & 512 & 1024 & 2048 & 4096 \\
  \hline
  SimSiam    & 68.1 & 68.0 & 67.9 & 64.0 \\

  VICReg              & 68.2 & 68.3 & 68.6 & 67.8 \\
  IMSVD               & \textbf{68.3} & \textbf{69.3} & \textbf{69.4} & \textbf{68.7} \\
    \hlineB{2}
  \end{tabular}
  \label{tab:batchsize}
\end{table}

\subsubsection{Effect of Batch Size.}
SSL methods usually require a large batch size, especially for contrastive learning.
Here we evaluated IMSVD with different batch sizes and the results are reported in Table~\ref{tab:batchsize}.
It shows that IMSVD achieved consistently better results than VICReg over different batch sizes.
As discussed in Sec.~\ref{sec_analysis}, an intrinsic property of IMSVD is to discriminatively encode different instances, making it work well without a large number of contrastive samples.

\begin{table}[htp]
  \renewcommand\tabcolsep{18pt}
 \caption{\textbf{Projector Depth}. The best results are highlighted in \textbf{bold}.}
  \centering
  \begin{tabular}{cccc}
  \hlineB{2}
         Depth         & 2 & 3  & 4 \\
        \hline
         Top-1               & \textbf{69.4} & 68.5 & 67.9 \\
         Top-5               & \textbf{89.3} & 88.3 & 87.9 \\
        \hline
        Time/100ep               & 8.5h & 9.6h & 10.8h \\
        Memory/GPU               & 10.4G & 11.5G & 12.5G \\
        \hlineB{2}
  \end{tabular}
  \label{tab:projector}
\end{table}

\begin{table}[htp]
  \renewcommand\tabcolsep{9.5pt}
 \caption{\textbf{Feature Dimension}. The best Top-1 accuracies are highlighted in \textbf{bold}. }
  \centering
  \begin{tabular}{cccccc}
         \hlineB{2}
             $D_{\text{VICReg}}$         & 1024 & 2048 & 4096 & 8192  & 16384 \\
             $D_{\text{IMSVD}}$         & 960 & 2000 & 4080 & 8160  & 16320 \\
             \hline
             VICReg            &   62.4 & 65.1 & 67.3 &68.6    & 68.8  \\
             IMSVD              &   \textbf{64.1} & \textbf{66.6} & \textbf{69.2} & \textbf{69.4} & \textbf{69.1}  \\
             \hline
             Time/100ep &   7.6h & 7.7h & 8.0h & 8.5h    & 10.9h  \\
             Memory/GPU &   7.6G & 8.0G & 8.5G &10.4G    & 15.9G  \\
            \hlineB{2}
  \end{tabular}
  \label{tab:dim}
\end{table}

\begin{table}[htp]
  \renewcommand\tabcolsep{10pt}
 \caption{\textbf{Ablation study on loss terms}. The best results are highlighted in \textbf{bold}.}
  \centering
  \begin{tabular}{ccccc}
         \hlineB{2}
         Loss         & DE+OE  & OE+TI & DE+OE+TIC &  DE+OE+TI \\
        \hline
         Top-1               &  65.4  & 64.1 & 68.3 & \textbf{69.4} \\
         Top-5               &  86.9  & 86.4 & 88.6 & \textbf{89.3} \\
        \hlineB{2}
  \end{tabular}
  \label{tab:loss}
\end{table}

\begin{table}[htp]
  \renewcommand\tabcolsep{13pt}
 \caption{\textbf{Feature Dimension}. The best Top-1 accuracies are highlighted in \textbf{bold}. }
  \centering
  \begin{tabular}{cccccc}
         \hlineB{2}
         $D_M$         & 32 & 64 & 80 & 96 & 128  \\
        \hline
         Top-1            & 67.8 & 69.1 & \textbf{69.4}  &69.2 & 68.4 \\
         Top-5            & 88.5 & 89.1 & \textbf{89.3}  &89.1 & 88.5 \\
        \hlineB{2}
  \end{tabular}
  \label{tab:seg}
\end{table}

\subsubsection{Effect of Projector Depth}
The existing studies~\cite{Chen_2021_CVPR, zbontar2021barlow, bardes2021vicreg} show that using a three-layer MLP as the projector achieved the best results.
However, IMSVD has a different behavior that a two-layer MLP achieved the best results as shown in Table~\ref{tab:projector}.
It may be because IMSVD learns information-maximized embedding features with stronger representation ability, so a deeper extra projector is not necessary.
Moreover, the computational cost can be reduced, especially for the fully connected MLP with high-dimensional inputs and outputs.
The running time per 100 epochs and the peak memory per GPU for different projector depths are reported in Table~\ref{tab:projector}, where the computational environment is described in Table~\ref{sec_implement}.
The comparison results in Table~\ref{tab:run} show that IMSVD cannot only reduce the running time and memory cost but also achieve better performance than BT and VICReg.

\subsubsection{Effect of Feature Dimension}
In the previous BT and VICReg studies, it was found that a very high-dimensional embedding vector is necessary for improving the representation learning performance. For IMSVD, the feature dimension plays an important role as well. The results of different feature dimensions for VICReg and IMSVD are reported in Table~\ref{tab:dim}, where the dimensions of IMSVD embeddings are similar to those of VICReg embeddings while keeping the dimension of each one-hot sub-vector the same, \textit{$D_M = 80$}.
IMSVD achieved consistently better results than VICReg on different embedding feature dimensions.
When the embedding feature dimension was reasonably large (4,096 and 8,192), IMSVD achieves the best results that are even better than the best results of VICReg using the larger dimension of 16,384. 
This is because minimizing linear correlation by the existing methods cannot ensure the minimized non-linear dependency while IMSVD can minimize any form of dependency between any two feature variables.
The large embedding feature dimension (i.e., 16,384) significantly increases the computational and memory cost for the SSL methods that compute the covariance or joint probability matrix~\cite{zbontar2021barlow}.
This point is demonstrated in Table~\ref{tab:dim} by evaluating running time and memory cost, where the computational environment is described in Sec.~\ref{sec_implement}.

\label{sec_vis}
\begin{figure}[h]
    	\centering
		\includegraphics[width=0.5\textwidth]{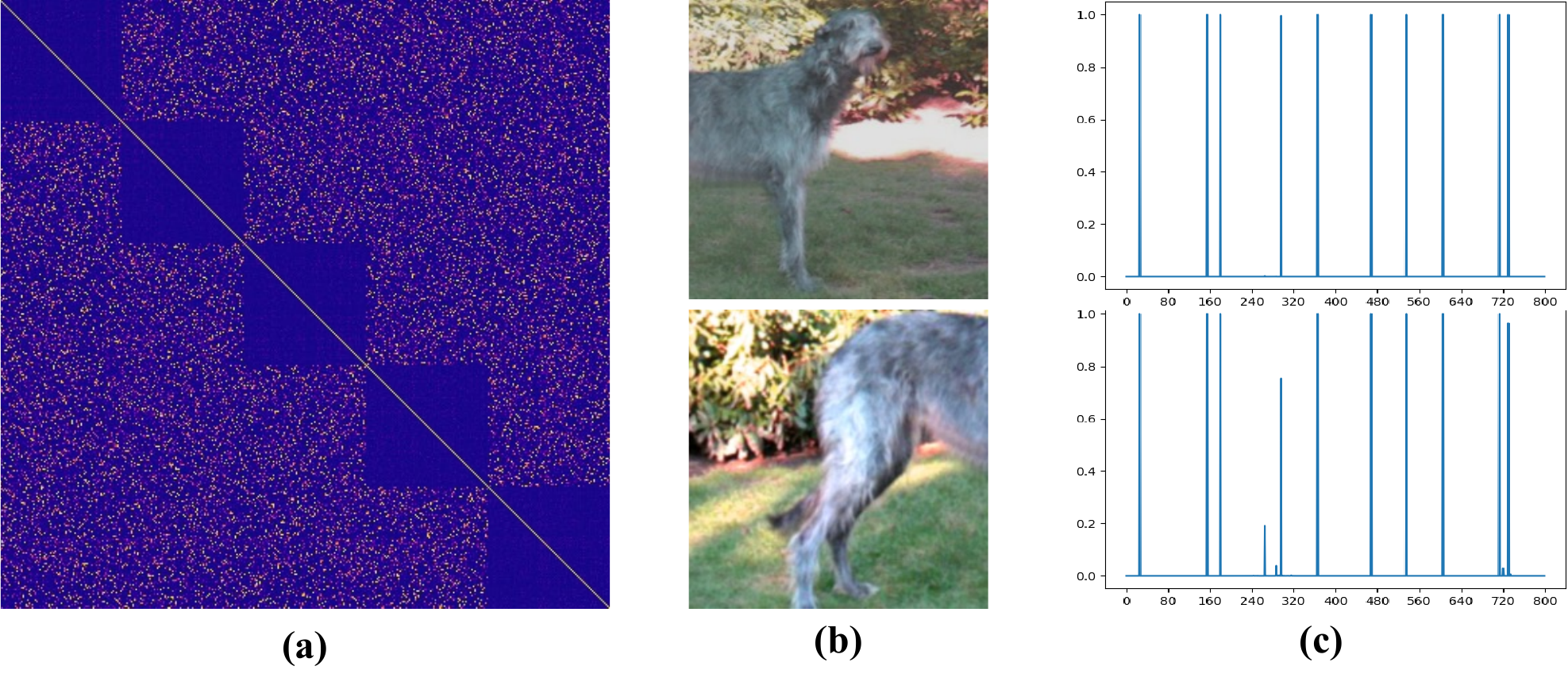}
        \caption{Visualization of IMSVD. (a) Cross-joint probability matrix, where blue and yellow respectively represent small and large values, and only the first five variables are shown for clear visualization. Note that this is computed on the whole ImageNet train set. (b) two transformations of the same image, and (c) Embedding vectors corresponding to the images in (b), where only the first ten variables are shown for clear visualization. Although we only show a single case for (b) and (c), readers can check more cases using our provided codes and models.}
		\label{fig:vis}
\end{figure}

\begin{figure*}[h]
    	\centering
		\includegraphics[width=0.9\textwidth]{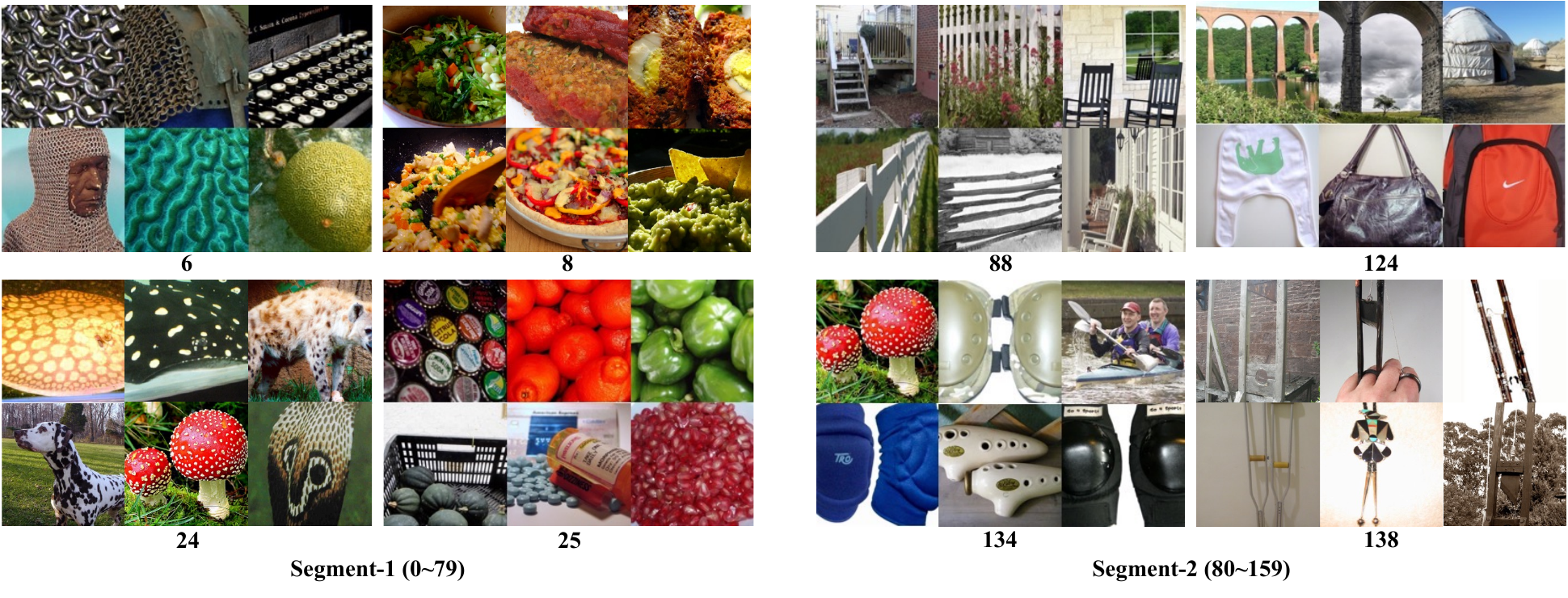}
        \caption{Visualization of learned IMSVD features on ImageNet validation set. The left side shows the samples assigned to the features indexed by 6, 8, 24, and 25 of the first variable. The right side shows the samples assigned to the features indexed by 88, 124, 134, and 138 of the second variable. Although we only show several cases, readers can check more cases using our provided codes and models.}
		\label{fig:samples}
\end{figure*}
\begin{figure*}[bt]
    \centering
    \includegraphics[width=0.9\textwidth]{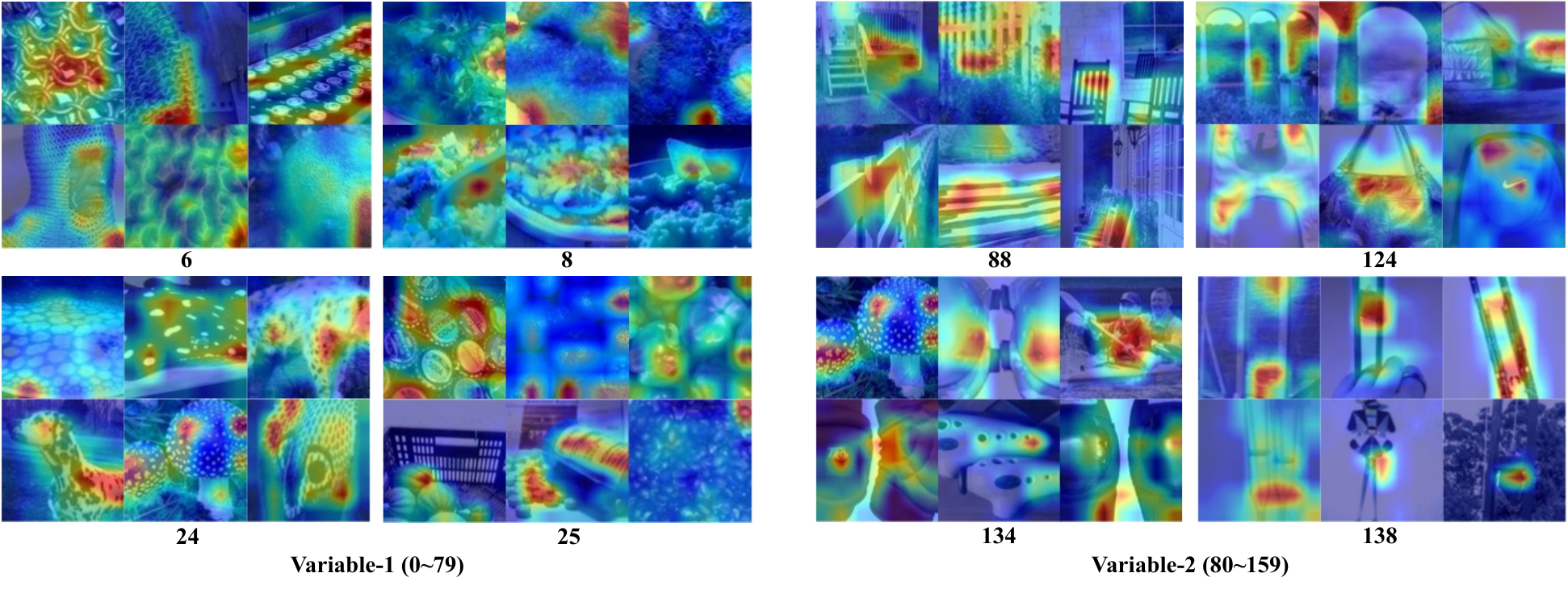}
    \caption{Local feature visualization using Grad-CAM. Here the Grad-CAM maps show the local features of the learned variables, where the same set of images are used as in Fig.~\ref{fig:samples}.}
    \label{fig:loc}
\end{figure*}

\subsubsection{Effect of Loss Function}
The effect of different loss terms was evaluated in Table~\ref{tab:loss}, where DE, OE, TIC, and TI denote the diagonal entropy loss, off-diagonal entropy loss, transformation invariance loss implemented with cross-entropy, and transformation invariance loss implemented with inner-product, respectively.
As described in the Appendix-II, only optimizing the entropy loss (DE+OE) allows IMSVD to avoid collapsed solutions and learn informational representations. 
This theoretical analysis is consistent with the empirical results in Table~\ref{tab:loss} that 65.4\% Top-1 was achieved using the joint-entropy loss only, comparable to some methods reported in Table~\ref{tab:compare}.
Adding the enhanced transformation invariance constraint at the instance level significantly improved the performance, as also discussed in Sec.~\ref{sec_method_loss}.
Without adding the DE loss, the results were significantly degraded, as the DE loss not only enhances the transformation invariance but also maximizes the information/entropy of each variable by enforcing a uniform distribution.
Minimizing the cross-entropy degraded the performance compared with the inner-product implementation.

\subsubsection{Effect of One-hot Sub-Vector Dimension}
Finally, the effect of our unique hyperparameter, \textit{i.e.}, the dimension of the one-hot sub-vector, was evaluated. Our empirical results with different dimensions in Table~\ref{tab:seg} indicate that $D_M = 80$ achieved the best results, where the dimension of the whole embedding vector was kept the same. It shows that the performance is not very sensitive to this hyperparameter.

\subsection{Visualization of IMSVD Features}
In Fig.~\ref{fig:vis}, we visualize IMSVD including an empirical joint probability matrix and individual embeddings, where the empirical joint probabilities were computed over the whole ImageNet train dataset and only the left-upper partial matrix of $400 \times 400$  with the first 5 variables and the individual variables with the first 800 units or the first 10 variables are selected for visualization. 
The theoretical analysis in Sec.~\ref{sec_analysis} demonstrates that the embedding statistics are enforced to be uniform; \textit{i.e.}, $\forall m, d, \mP(m, m, d, d)=\frac{1}{D_M}$, meaning that the probabilities of the diagonal elements in all diagonal blocks are equal and those of off-diagonal elements in the diagonal blocks are zeros.
Also, the probabilities of all elements of the off-diagonal blocks are equal, \textit{i.e.}, $\forall m_1, m_2, d_1, d_2, m_1 \ne m_2, \mP(m_1, m_2, d_1, d_2) = \frac{1}{(D_M)^2}$.
The empirical joint probability matrix visualized in Fig.~\ref{fig:vis}(a) is consistent with theoretical analysis although not a perfect match.
Furthermore, Figs.~\ref{fig:vis}(b) and (c) show that the embedding features of each variable tend to be one-hot and invariant to the transformations. 
Statistically, we computed all the discrete features with the trained model on ImageNet. Our results show that 91.18\% of the soft discrete sub-vectors have the highest value larger than 0.9, and 81.25\% of the soft discrete sub-vectors have the highest value larger than 0.99, indicating the feature variables are nearly one-hot encoded. All these empirical results are also consistent with the theoretical analysis.

Previous studies~\cite{kotsiantis2006discretization,Garca2013ASO} found that discretization can improve the interpretability.
To qualitatively evaluate if meaningful embedding features are learned by our IMSVD, in Fig.~\ref{fig:samples} we show some examples assigned to specific units of the first two variables, where the whole embedding vector has 8,160 units including 102 variables, and each variable has 80 units.
Specifically, some features of the first variable represent different types of textures; \textit{e.g.}, the feature unit indexed by 24 represents the dot style textures, and the units 6, 8, and 25 correspond to other specific textures/patterns. Some features of the second variable represent different shapes; \textit{e.g.}, unit 124 abstracts a ``$\cap$'' shape, and units 88, 134, 138 represent other shapes/patterns.
Obviously, the first and second variables use different principles to group samples; \textit{e.g.}, the first variable groups the image containing red mushrooms (indexed by 24) with the objects having similar textures, while the second variable groups it (indexed by 134) with the images having twin/repeated objects. Since each sub-vector can be regarded as an unsupervised cluster/classification head~\cite{niu2020gatcluster, spice},  the Grad-CAM~\cite{selvaraju2017grad} algorithm can be easily implemented to visualize the corresponding local regions to the learned variables in Fig.~\ref{fig:loc}. It can be seen that the above-described local features are well-captured by learned feature variables although not perfect.
These visual results indicate that some discrete variables are indeed explainable, although we cannot ensure that all discrete units correspond to the well-known attributes.

\section{Conclusion}

We presented an Information-Maximized Soft Variable Discretization (IMSVD) method that softly discretizes latent variables and enables a new information-theoretic objective for self-supervised image representation learning.
Theoretical analysis ensures that the optimized IMSVD embedding features are discrete, transform-invariant, non-collapsed, redundancy-minimized, and discriminative. IMSVD can minimize any form of dependency between feature variables beyond the linear correlation in current methods. We show that our non-contrastive IMSVD method actually performs contrastive learning in an information-maximized way.
Experimental results have shown the effectiveness and superiority of IMSVD in terms of both accuracy and efficiency.
Like other hyperparameters (e.g., learning rate), the introduced hyperparameter ($D_M$) needs to be empirically selected on different datasets. Future work is needed to automatically identify this hyperparameter in a principled way.
Nevertheless, IMSVD has the potential to be adapted for more tasks and other learning paradigms, such as hierarchical clustering and multi-modality feature alignment.


 
\bibliographystyle{IEEEtran}
\bibliography{IEEEabrv,./reference}

\vfill

\end{document}

%% file: math_commands2.tex
\usepackage{amsmath,amsfonts,bm}

\def\1{\bm{1}}

\def\vtheta{{\bm{\theta}}}

\def\vd{{\bm{d}}}

\def\vp{{\bm{p}}}
\def\vq{{\bm{q}}}

\def\vx{{\bm{x}}}

\def\vz{{\bm{z}}}

\def\mP{{\bm{P}}}

\DeclareMathAlphabet{\mathsfit}{\encodingdefault}{\sfdefault}{m}{sl}
\SetMathAlphabet{\mathsfit}{bold}{\encodingdefault}{\sfdefault}{bx}{n}

\def\sX{{\mathbb{X}}}

\newcommand{\R}{\mathbb{R}}

